\documentclass{sbrt}
\usepackage[english,brazil]{babel}
\usepackage[utf8]{inputenc}

\usepackage{graphicx}
\graphicspath{ {./images/} }

\usepackage{amsmath}

\usepackage{multirow}
\usepackage{cite}

\usepackage{booktabs}
\usepackage[table]{xcolor}

\begin{document}

\title{Determinação Automática de Limiar de Detecção de Ataques em Redes de Computadores Utilizando Autoencoders}

\author{Luan Gonçalves Miranda, Pedro Ivo da Cruz e Murilo Bellezoni Loiola
\thanks{Luan Gonçalves Miranda, CECS, UFABC, Santo André-SP, e-mail: luan.miranda@ufabc.edu.br; Pedro Ivo da Cruz, CECS, UFABC, Santo André-SP, e-mail: pedro.cruz@ufabc.edu.br; Murilo Bellezoni Loiola, CECS, UFABC, Santo André-SP, e-mail: murilo.loiola@ufabc.edu.br. Este trabalho foi parcialmente financiado pela CAPES (001).}%
}

\maketitle

\markboth{XL SIMPÓSIO BRASILEIRO DE TELECOMUNICA\c{C}\~{O}ES E PROCESSAMENTO DE SINAIS - SBrT 2022, 25--28 DE SETEMBRO DE 2022, STA. RITA DO SAPUCAÍ, MG}{}

\begin{resumo}
Atualmente, mecanismos de segurança digital, como os sistemas de detecção por anomalia utilizando \textit{Autoencoders} (AE) mostram grande potencial por desviar de problemas intrínsecos aos dados como, por exemplo, o desbalanceamento. Pelo fato dos AE utilizarem um limiar de separação, de definição não trivial e não padronizada, para classificar o erro de reconstrução extraído, a definição deste limiar impacta diretamente o desempenho do sistema de detecção. Portanto, este trabalho propõe a definição automática deste limiar utilizando alguns algoritmos de aprendizado de máquina. Para tanto, foram avaliados três algoritmos: o \textit{K-Nearst Neighbors}, o \textit{K-Means} e a \textit{Support Vector Machine}.
\end{resumo}
\begin{chave}
Detecção de Anomalia, Auto Codificadores, Ataques em Redes.
\end{chave}

\begin{abstract}
Currently, digital security mechanisms like Anomaly Detection Systems using Autoencoders (AE) show great potential for bypassing problems intrinsic to the data, such as data imbalance. Because AE use a non-trivial and non-standardized separation threshold to classify the extracted reconstruction error, the definition of this threshold directly impacts the performance of the detection process. Thus, this work proposes the automatic definition of this threshold using some machine learning algorithms. For this, three algorithms were evaluated: the K-Nearst Neighbors, the K-Means and the Support Vector Machine.
\end{abstract}

\section{Introdu\c{c}\~{a}o}

Atualmente, a segurança digital pode ser vista como uma área de extrema importância econômica, tendo em vista o ativo de altíssimo valor que se tornou a informação, denominando a época atual como a "Era da Informação". Com isso, mecanismos de proteção de dados e sistemas vêm sendo desenvolvidos para suprir esta necessidade. Uma das abordagens estudadas é a dos Sistemas de Detecção de Intrusão (IDS, do inglês \textit{Intrusion Detection Systems}), responsáveis por monitorar e inspecionar as atividades em rede com o intuito de detectar ameaças que tentem violar a segurança~\cite{Moustafa2019, Ahmed2016}.

A estrutura clássica de um IDS baseia-se em quatro componentes: o Decodificador, que traduz os dados de tráfego codificados pelo protocolo IP e tem como missão decodificar estes dados para serem utilizados posteriormente; o Pré-processador, onde é feita a extração, criação, redução, conversão ou normalização dos dados decodificados; o Sistema de Decisão, onde se realiza a detecção de ataque; e o Mecanismo de Defesa, sendo executado ao detectar um ataque~\cite{Moustafa2019, Ahmed2016}.

Em um IDS, o componente mais importante é o sistema de decisão, pois é nele que todo o processo de detecção de ataque ocorre. Duas abordagens podem ser utilizadas em sua construção: a abordagem por Mau-Uso, dando origem aos Sistemas de Detecção por Mau-Uso (MDS, do inglês \textit{Misuse Detection Systems}) e a abordagem por Anomalia, dando origem aos Sistemas de Detecção por Anomalia (ADS, do inglês \textit{Anomaly Detection Systems})~\cite{Fernandes2019, Kurniabudi2019}. Nos MDS, são utilizadas características previamente conhecidas dos ataques para realizar a detecção dos mesmos. Já nos ADS, é utilizado o desvio de padrão nos dados, sendo este padrão determinado pelo uso normal da rede. A hipótese é de que um ataque, ou uso anômalo da rede, provocará alteração nos dados coletados pelo IDS, permitindo assim a sua identificação. Uma vantagem dos ADS sobre os MDS é a não necessidade de conhecimento específico dos ataques a serem detectados~\cite{Fernandes2019, Kurniabudi2019}.

Apesar das diferenças entre as características dos dados de usuários normais e maliciosos serem pequenas, o uso de técnicas específicas, como as técnicas de aprendizado de máquina, possibilita a realização desta distinção. Em~\cite{Manna2019}, por exemplo, três técnicas foram investigadas para fazer a detecção de um ataque: \textit{Random Forests}, árvore de decisão e \textit{REPTree}. O trabalho utilizou uma base de dados SNMP-MIB para treinar e avaliar as técnicas. Esta mesma base de dados também foi utilizada em~\cite{Al-Naymat2018}, que avaliou o \textit{AdaboostM1}, uma \textit{RandomForest} e uma Rede Neural Perceptron Multicamadas na detecção de ataques. Já em~\cite{Toupas2019} foi avaliado o uso de Redes Neurais Profundas, na base de dados CICIDS2017.

Embora amplamente utilizados em problemas de classificação, estes algoritmos não apresentam bom desempenho quando os dados disponíveis para treinamento são desbalanceados, isto é, quando há uma predominância considerável de determinadas classes em relação a outras. Este é um problema comum em IDS, visto que os dados de ataques, normalmente, aparecem em menor proporção. Neste sentido, os Auto Codificadores (AE, do inglês \textit{Autoencoders}) têm sido utilizados por permitir o treinamento usando apenas a classe mais comum~\cite{Charte2018, Sakurada2014}. Assim, o AE reduzirá o Erro de Reconstrução (RE, do inglês \textit{Reconstruction Error}) apenas desta classe dominante, sendo que uma classe anômala produzirá um RE de maior amplitude, permitindo a distinção entre às duas classes.

Esta característica de treinamento de AE possibilita sua utilização como um classificador através do RE extraído dele, sendo RE considerado pequeno para a classe de treinamento e maior para qualquer outra. A decisão sobre qual classe determinado RE pertence é feita através da sua comparação com um limiar. Em geral, este limiar é calculado utilizando testes de hipótese, resultando em diferentes critérios para o seu cálculo~\cite{Madani2018, Zavrak2020, Choi2019}.
Alguns trabalhos, como os dos artigos~\cite{Madani2018, Zavrak2020, Choi2019}, que utilizam o RE para realizar a detecção de anomalia, utilizam formas diferentes de definição desse limiar. Em~\cite{Madani2018}, foi utilizado um AE eliminador de ruído e também o algoritmo de Análise de Componentes Principais (PCA, do inglês \textit{Principal Component Analysis}). O limiar de separação foi obtido de forma a minimizar a taxa de falso positivos, i.e., a taxa do sistema identificar um ataque quando este não está ocorrendo. Em~\cite{Zavrak2020}, foram utilizadas duas classes de AE, o AE básico e o AE Variacional, comparadas a uma Máquina de Vetor Suporte (SVM, do inglês \textit{Support Vector Machine}). O limiar não é explicitamente calculado. Ao invés disto, vários valores de limiares diferentes foram escolhidos arbitrariamente e o desempenho das técnicas foram avaliadas também em função da escolha do limiar. Já em~\cite{Choi2019},  foi considerado um ADS utilizando AE quando havia Contaminação Adversária (AC, do inglês \textit{Adversarial Contamination}), ou seja, quando existe, nos dados normais utilizados no treinamento do AE, uma porção de dados anômalos. Isto prejudica o desempenho do ADS. Para uma análise mais profunda, proporções de AC foram consideradas no trabalho. Também foi estabelecido um critério para o cálculo do limiar considerando esta proporção. No entanto, esta proporção nem sempre é previamente conhecida, dificultando a obtenção do limiar em uma situação mais prática.

Como visto, a obtenção de um limiar para conseguir fazer a detecção utilizando AE continua sendo uma questão em aberto. Neste trabalho, ao invés de realizar o cálculo explícito do limiar, que depende muitas vezes de variáveis desconhecidas pelo projetista do sistema, propomos que esta decisão seja tomada por outro algoritmo de aprendizado de máquina. Foram considerados os algoritmos K-Vizinhos Mais Próximos (KNN, do inglês \textit{K-Nearest Neighbors}), K-means e SVM para realizar esta tarefa. Como estes algoritmos também precisam passar por uma etapa de treinamento, também é proposto um \textit{pipeline} para o treinamento do sistema completo, contendo o AE  e o algoritmo de decisão.

O restante do trabalho é organizado da seguinte forma: a Seção II descreve as técnicas utilizadas e a Seção III detalha as etapas e ferramentas propostas neste trabalho. Já a Seção IV apresenta e discute os resultados obtidos e, por fim, a Seção V conclui o trabalho.

\section{Fundamentação Teórica}
\label{Embasamento Teórico}

\subsection{Autoencoders}
AE são redes neurais que, diferente do objetivo clássico de classificação de dados, visam reconstruir na saída o sinal de entrada após este passar por camadas intermediárias de compressão e descompressão. Estas camadas de compressão consistem em camadas com menos neurônios que as de entrada e de saída~\cite{Charte2018}. Com isso, a particularidade dos AE se dá pela estrutura da rede neural utilizada, onde a rede apresenta um gargalo em sua Camada Intermediária (IL, do inglês \textit{Intermediate Layer}), representando uma compressão dos dados. A estrutura consiste em dois blocos, o codificador (do inglês \textit{Encoder}) e o decodificador (do inglês \textit{Decoder}). Codificador refere-se às camadas iniciais até a IL da rede, tendo a função de comprimir os dados de entrada. Já o decodificador refere-se às camadas finais da rede, objetivando à descompressão dos dados comprimidos e, consequentemente, à reconstrução dos dados da entrada. Estas etapas de compressão e descompressão só são possíveis pelo fato de existirem correlações entre as características inseridas na entrada~\cite{Charte2018}.

\begin{figure}
    \centering
    \includegraphics[scale=0.4]{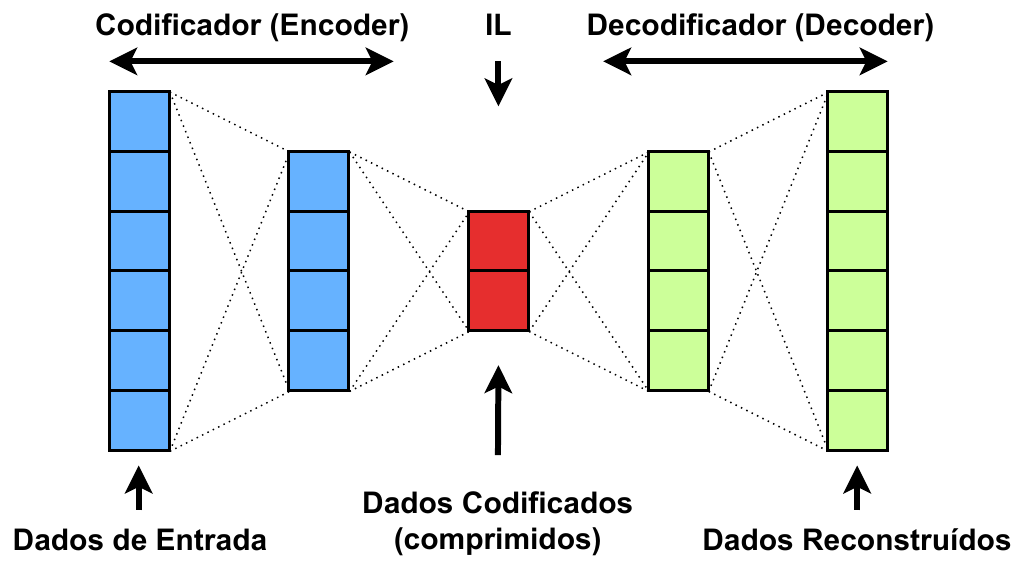}
    \caption{Exemplo de Representação de um AE com duas camadas no codificador e decodificador.}
    \label{fig:Autoencoder}
\end{figure}

Um exemplo de um AE pode ser visualizado na Fig.~\ref{fig:Autoencoder}. Neste exemplo, o AE é composto por 5 camadas: duas de codificação (neurônios azuis), uma IL (neurônios vermelhos) e duas de decodificação (neurônios verdes).

Duas aplicações diretas dos AE são a redução de dimensionalidade e a remoção de ruído dos dados~\cite{Charte2018, Sakurada2014}, sendo uma alternativa ao PCA~\cite{Sakurada2014}, algoritmo clássico utilizado para redução de dimensionalidade, uma vez que realiza mapeamentos não lineares, podendo fazer projeções mais adequadas dos dados~\cite{Hwang1999}. 
Uma aplicação diferente dos AE  refere-se a sua utilização para classificação de dados através da utilização de alguns parâmetros extraídos e utilizados como características, como no caso do RE~\cite{Madani2018, Zavrak2020, Yu2017}. Apesar da eficiência da reconstrução dos dados, a compressão existente na IL da rede provoca um pequeno RE. A tentativa de construção de um AE eficiente consiste em reduzir este erro de tal forma que a saída seja a mais próxima possível à entrada. Esta forma de utilizá-lo possibilita a construção de ADS baseado em AE.

\subsection{Autoencoders em ADS}
AE podem ser utilizados em ADS por permitirem a extração de algumas características, sendo a principal delas o RE. A premissa por trás da utilização do RE tem a ver com o fato de que, dado que o AE foi treinado para minimizar o RE com dados da classe majoritária, o RE desta classe sempre estará próximo a zero. Por outro lado, ao receber dados anômalos na sua entrada, o RE resultante será maior. Com isso, o RE é, portanto, uma característica extraída pelo AE e que pode ser utilizada como base para decidir se aqueles dados na entrada correspondem a um ataque ou não.

A principal questão, nesta categoria de utilização, é a escolha eficiente do limiar de separação, como visto anteriormente. O presente trabalho propõe a utilização de alguns algoritmos de aprendizado de máquina para tomar esta decisão, sendo realizada abaixo uma breve descrição desses algoritmos utilizados.

\subsubsection{Limiar de Referência}
Como base de comparação, utilizou-se o método de cálculo do limiar proposto em~\cite{Choi2019}, onde foi utilizada a equação da Pontuação Z (ZS, do inglês \textit{Z-Score}) para definição do limiar $\theta$, sendo dada por 
\begin{equation}
    \theta = \mu_{RE} + Z_{AC} * \sigma_{RE}
\end{equation}
sendo $\mu_{RE}$ a média e $\sigma_{RE}$ o desvio padrão dos REs obtidos, e o $Z_{AC}$ o valor do \textit{ZS} relacionado com a porcentagem de AC no treinamento. Como aqui não está sendo utilizada AC, o limiar resulta apenas na média do RE. Com isso, para a classificação dos dados entre normal e anômalo, a seguinte equação foi utilizada:
\begin{equation}
\text{Referência} = \left\{\begin{array}{ccc}
    \text{normal} & \text{se} & RE \leq \mu_{RE}\\
    \text{anomalo} & \text{se} & RE > \mu_{RE}\\
\end{array}\right.
\end{equation}

\subsubsection{K-Nearest Neighbors}

O algoritmo KNN~\cite{Kramer2013} é um dos clássicos algoritmos de aprendizado supervisionados. A classificação do KNN é realizada através da similaridade entre dados, ou melhor, a partir da distância entre uma instância de dado e seus vizinhos. Existem diversas formas de calcular esta distância, sendo a mais comum a Distância Euclidiana (ED, do inglês \textit{Euclidean Distance}).

\subsubsection{KMeans}
O algoritmo \textit{K-Means}\cite{Blomer2016} é um algoritmo que faz parte da classe de técnicas não-supervisionada, ou seja, que não necessitam de apresentação das classes dos dados em seu treinamento. Este algoritmo é baseado na formação de grupos (do inglês \textit{clusters}) dos dados em torno de um centroide.

\subsubsection{SVM}
SVM~\cite{Yu2008} é um algoritmo supervisionado baseado na teoria de aprendizado estatístico, cujo objetivo é encontrar um hiperplano ideal que separe as classes de dados diferentes. Para isso, são utilizados os chamados Vetores de Suportes, sendo eles os dados localizados nas extremidades das classes. Com isso, é tomado o hiperplano na média destes vetores.

\section{Técnica Proposta}
Para o objetivo deste trabalho, foram extraídos dois conjuntos de características do AE: o RE e as valores da Saída da IL (SIL). Com isso, três modelos de ADS foram avaliados: um utilizando como característica somente o RE, outro utilizando somente a SIL e, por fim, um modelo onde a SIL e o RE foram utilizados em conjunto como característica do ADS. A SIL foi utilizada, diferente dos trabalhos mostrados anteriormente, para avaliação da codificação do AE como característica para detecção em um ADS.

\begin{figure*}
    \centering
    \includegraphics[scale=0.7]{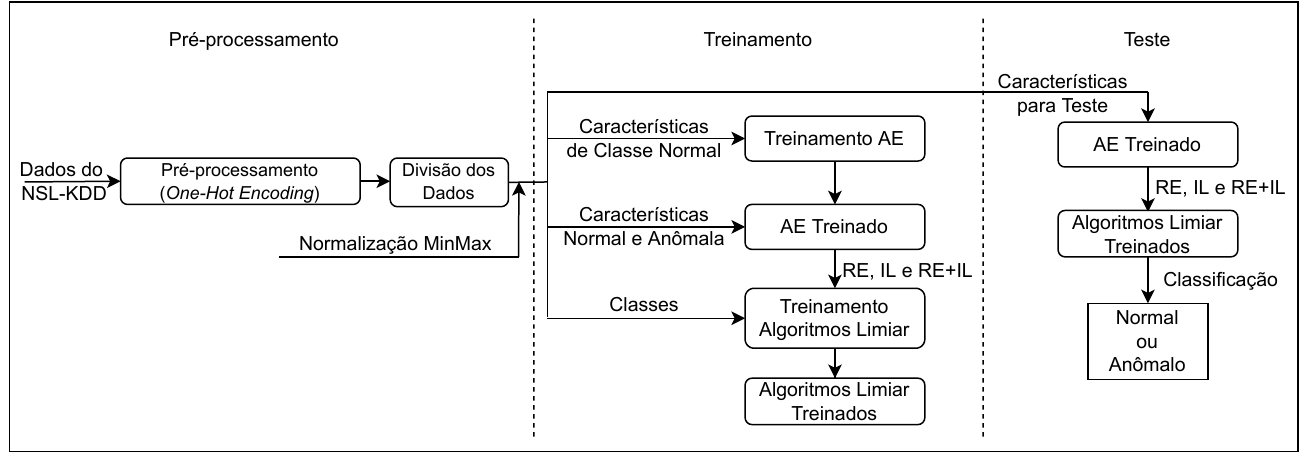}
    \caption{Fluxo das etapas da técnica proposta.}
    \label{fig:Fluxo}
\end{figure*}

Na Fig.~\ref{fig:Fluxo}, é possível visualizar o fluxo da técnica aqui proposta, sendo detalhada cada etapa nas subseções seguintes.

\subsection{Conjunto de Dados}
Para avaliação da técnica proposta, foi utilizado o conjunto de dados NSL-KDD (NSL-KDD, do inglês \textit{National Security Laboratory - Knowledge Discovery and Data Mining Tools})~\cite{NSL-KDD}, que é uma derivação do conjunto desenvolvido em 1999 para a competição de Descoberta de conhecimento e Ferramentas de Mineração de Dados (KDD, do inglês \textit{Knowledge Discovery and Data Mining Tools})~\cite{KDD99}, corrigindo erros e informações redundantes.


\subsection{Pré-processamento}

O conjunto de dados NSL-KDD, já mencionado, é dividido em duas bases diferentes: uma para treinamento e outra para teste. Porém, como o método de treinamento utilizado aqui difere desta divisão pré realizada, inicialmente esses dois conjuntos de dados foram concatenados e, em seguida, foi realizada uma divisão mais adequada para a técnica aqui proposta, que se baseiam em três bases diferentes: uma para treinamento do AE, outra para treinamento dos algoritmos de definição de limiar e outra para teste. Pela utilização do método \textit{k-fold} de validação cruzada na avaliação, novamente o conjunto é dividido em dois: um para treinamento (redividido para treino do AE e do algoritmo de limiar) e outro para teste. Baseado na quantidade de dados normais presentes no conjunto de treinamento de cada \textit{fold}, foi realizada uma divisão aleatória de 50\% destes dados para treinamento do AE, e os outros 50\% para treinamento dos algoritmos de detecção do limiar, sendo este último balanceado com dados anômalos, tendo em vista a necessidade de mais de uma classe no treino desses algoritmos, diferentemente do AE.

Antes da divisão dos dados, a variável \textit{service} foi retirada dos conjuntos de dados por existirem categorias na base de treinamento que não existiam na base de teste. Além disso, é realizado o pré-processamento \textit{One-Hot Encoding}, onde as variáveis nominais são transformadas em variáveis numéricas.

Por fim, a normalização \textit{MinMax} foi realizada em cada uma das três bases obtidas, tendo sido utilizado o mesmo fator de normalização para todas as bases, sendo este fator obtido a partir da base de treinamento dos algoritmos de definição do limiar.

\subsection{Treinamento e Teste}
O treinamento do modelo proposto neste trabalho é realizado em etapas. Inicialmente, o AE foi treinado utilizando o conjunto apropriado obtido no pré-processamento. Posteriormente, foi aplicado na entrada do AE o conjunto de dados de treinamento dos algoritmos de definição do limiar, sendo extraído o RE e/ou a SIL, utilizados como características. Estas características foram utilizadas para treinamento dos algoritmos (KNN, \textit{K-Means} e SVM) finalizando o treinamento do modelo.

Com o modelo treinado, é seguido o mesmo fluxo dos dados no treinamento, porém utilizando o conjunto de teste. Com isso, os dados são inseridos na entrada do AE, seguido da extração do RE e/ou da SIL, sendo estes dados inseridos nos algoritmos para classificação entre dado Normal ou Anômalo.


\subsection{Estrutura dos Algoritmos}
O AE utilizado possuía 5 camadas, sendo 2 de codificação, a IL, e 2 de decodificação, sendo esta escolha de estrutura realizada de forma aleatória tendo em vista que esta definição é algo subjetivo no projeto de redes neurais. Porém, foi seguida a regra implícita de definição de neurônios em cada camada do AE, onde as camadas externas devem sempre ter uma quantidade maior de neurônios do que as camadas internas, representando a compressão dos dados. Com isso, considerando \textit{N} como a quantidade de características do conjunto de dados utilizado, sendo \textit{N} maior que o valor das camadas internas, a quantidade de neurônios de cada camada foi igual a \textit{N}, 32, 16, 32 e \textit{N}, seguindo da camada inicial até a final. Além disso, foi utilizada como função de ativação a função \textit{relu} em todas as camadas.

Nas estruturas dos algoritmos utilizados para definição do limiar foram utilizados 11 vizinhos no KNN, 2 \textit{clusters} no \textit{K-Means} e como \textit{kernel} na SVM utilizou-se a função \textit{rbf}.

\section{Resultados e Avaliação}
Para desenvolvimento dos modelos foi utilizada a linguagem de programação \textit{Python} desenvolvido no editor \textit{Jupyter}, conjuntamente com as bibliotecas de aprendizado de máquina \textit{Keras}, \textit{TensorFlow} e \textit{Scikit-Learn}.

Após a implementação de todos os algoritmos, foram obtidos os resultados com a utilização das seguintes métricas: acurácia, precisão, \textit{recall} e \textit{f1-score}, sendo eles apresentados nas subseções seguintes. Os resultados mostrados correspondem às médias e desvios padrão obtidos através da validação cruzada com \textit{k-fold}, com $k=10$. Além disso, foram utilizadas 50 épocas de treinamento, tamanho de \textit{batch} de 256 e função de otimização \textit{Adam} com taxa de aprendizagem de 0,0001.

\subsection{Erro de Reconstrução}
\label{Erro de Reconstrução}

Utilizando somente o RE como característica extraída pelo AE e usada como entrada no algoritmo de classificação, os resultados obtidos são apresentados na Tabela~\ref{tabela1}. Pode-se observar que o KNN apresenta o melhor desempenho, apresentando apenas um valor de precisão inferior ao da Referência. Comparando as matrizes de confusão média, apresentadas na Tabela~\ref{tabela2}, pode-se ver que a Referência alcançou a média de 12550,2 acertos ($7368,3+5181,9$) e 2302,3 erros ($337+1965,3$), e o KNN 13478,7 acertos ($7159,6+6319,1$) e 1374,8 erros ($547,7+827,1$), tendo o KNN apresentado maior taxa de falsos positivos (quando a classe desejada é normal, mas é classificada como anômala) do que a referência, sendo esta taxa relacionada com a precisão. Este comportamento ocorre devido à pequena diferença entre as características dos usuários normais e anômalos, onde o RE de alguns dados podem destoar do RE padrão de sua classe, gerando confusão nos algoritmos de aprendizado de máquina, provocando o aumento de falsos negativos e positivos. Pode-se notar também que o KNN foi o mais estável ao comparar seu desvio padrão com os outros algoritmos.

O segundo melhor desempenho foi obtido pela SVM. No entanto, é possível observar o mesmo comportamento em relação à precisão. Em seguida, temos, respectivamente, a técnica da Referência e o \textit{K-Means}.

\begin{table}[htb]
\caption{\label{tabela1}Média dos resultados de cada fold para o uso somente do RE}
\begin{center}
{\tt
\resizebox{9cm}{!}{
\begin{tabular}{|c||c|c|c|c|}\hline
&Referência&KNN&K-Means&SVM\\\hline\hline
Acurácia & \begin{tabular}[c]{@{}c@{}}0.845 \\$\pm$ 0.109\end{tabular} & \begin{tabular}[c]{@{}c@{}}\textbf{0.907}\\ \textbf{$\pm$ 0.011}\end{tabular} & \begin{tabular}[c]{@{}c@{}}0.311\\ $\pm$ 0.280\end{tabular} & \begin{tabular}[c]{@{}c@{}}0.896\\ $\pm$ 0.014\end{tabular}\\\hline
Precisão & \begin{tabular}[c]{@{}c@{}}\textbf{0.956}\\ \textbf{$\pm$ 0.019}\end{tabular} & \begin{tabular}[c]{@{}c@{}}0.929\\  $\pm$  0.014\end{tabular} & \begin{tabular}[c]{@{}c@{}}0.221\\ $\pm$ 0.373\end{tabular} & \begin{tabular}[c]{@{}c@{}}0.926\\ $\pm$ 0.019\end{tabular}\\\hline
Recall & \begin{tabular}[c]{@{}c@{}}0.809\\  $\pm$  0.103\end{tabular} & \begin{tabular}[c]{@{}c@{}}\textbf{0.896}\\  \textbf{$\pm$  0.011}\end{tabular} & \begin{tabular}[c]{@{}c@{}}0.206\\ $\pm$ 0.300\end{tabular} & \begin{tabular}[c]{@{}c@{}}0.880\\ $\pm$ 0.020\end{tabular}\\\hline
F1-Score & \begin{tabular}[c]{@{}c@{}}0.872\\ $\pm$ 0.065\end{tabular} & \begin{tabular}[c]{@{}c@{}}\textbf{0.912}\\  \textbf{$\pm$  0.010}\end{tabular} & \begin{tabular}[c]{@{}c@{}}0.208\\ $\pm$ 0.333\end{tabular} & \begin{tabular}[c]{@{}c@{}}0.902\\ $\pm$  0.013\end{tabular}\\\hline
\end{tabular}
}
}
\end{center}
\end{table}

\begin{table}[htb]
\caption{\label{tabela2}Matrizes de Confusão Médias}
\begin{center}
{\tt
\begin{tabular}{|c|c||c|c|}\hline
\multicolumn{4}{|c|}{\textbf{Referência}}\\\hline
\multicolumn{2}{|c||}{\multirow{2}{*}{Classes}} & \multicolumn{2}{c|}{Classificada}\\\cline{3-4}
\multicolumn{2}{|c||}{} & \multicolumn{1}{l|}{\textbf{Normal}} & \textbf{Anômalo}\\\hline\hline
\multicolumn{1}{|l|}{\multirow{2}{*}{Desejada}} & \textbf{Normal} & \multicolumn{1}{l|}{7368,3}  & 337 \\\cline{2-4}
\multicolumn{1}{|l|}{} & \textbf{Anômalo} & \multicolumn{1}{l|}{1965,3} & 5181,9\\\hline\hline\hline

\multicolumn{4}{|c|}{\textbf{KNN}}\\\hline
\multicolumn{2}{|c||}{\multirow{2}{*}{Classes}} & \multicolumn{2}{c|}{Classificada}\\\cline{3-4}
\multicolumn{2}{|c||}{} & \multicolumn{1}{l|}{\textbf{Normal}} & \textbf{Anômalo}\\\hline\hline
\multicolumn{1}{|l|}{\multirow{2}{*}{Desejada}} & \textbf{Normal} & \multicolumn{1}{l|}{7159,6}  & 547,7 \\\cline{2-4}
\multicolumn{1}{|l|}{} & \textbf{Anômalo} & \multicolumn{1}{l|}{827,1} & 6319,1\\\hline
\end{tabular}
}
\end{center}
\end{table}

\subsection{Erro de Reconstrução e Camada Intermediária}
\label{Erro de Reconstrução e Camada Intermediária}
Dois outros cenários foram avaliados: utilizando como características na entrada do classificador (i) o RE em conjunto com a SIL do AE; (ii) somente a SIL do AE. Os resultados obtidos são apresentados na Tabela~\ref{tabela4}.

Nos dois cenários, o melhor resultado foi obtido com o KNN, seguido pela SVM e, por último, o \textit{K-Means}. Pode-se observar que ao se utilizar o RE em conjunto com a SIL não resultou em um melhor desempenho para o KNN e a SVM quando comparado com o caso que se utiliza somente a SIL.

Comparando os resultados obtidos para o \textit{K-Means} nos dois cenários, utilizando somente a SIL foram obtidos valores mais altos do que com a utilização do RE em conjunto com a SIL. Porém, nos dois casos observa-se um desvio padrão alto, indicando uma grande variabilidade no desempenho do algoritmo em relação a sua inicialização.

Comparando os resultados obtidos nas seções~\ref{Erro de Reconstrução} e~\ref{Erro de Reconstrução e Camada Intermediária}, nota-se que a utilização somente da SIL mostrou-se uma melhor opção, tendo em vista os resultados obtidos. Por exemplo, no KNN, todas as métricas ficaram em torno de 98\%. Essa melhora no desempenho, no entanto, ocorre em detrimento de um aumento da complexidade do classificador, visto que o uso dos dados da SIL possuem uma maior dimensionalidade do que somente o RE e, portanto, exigem um maior esforço para realizar o treinamento e a classificação dos dados.



\begin{table}[htb]
\caption{\label{tabela4}Médias dos resultados para uso do RE em conjunto com a SIL e somente a SIL}
\begin{center}
{\tt
\resizebox{9cm}{!}{
\begin{tabular}{|c||c|c|c|}\hline
\multicolumn{4}{|c|}{\textbf{RE e SIL}}\\\hline
&KNN&K-Means&SVM\\\hline\hline
Acurácia & \textbf{0.981  $\pm$  0.001} & 0.487 $\pm$ 0.325 & 0.960 $\pm$ 0.003\\\hline
Precisão & \textbf{0.981  $\pm$  0.001} & 0.495 $\pm$ 0.441 & 0.975 $\pm$ 0.003 \\\hline
Recall & \textbf{0.981  $\pm$  0.001} & 0.415 $\pm$ 0.346 & 0.949 $\pm$ 0.004\\\hline
F1-Score & \textbf{0.981  $\pm$  0.001} & 0.449 $\pm$ 0.388 & 0.962 $\pm$ 0.003\\\hline\hline\hline

\multicolumn{4}{|c|}{\textbf{SIL}}\\\hline
&KNN&K-Means&SVM\\\hline\hline
Acurácia & \textbf{0.980 $\pm$ 0.001} & 0.615 $\pm$ 0.294 & 0.957 $\pm$ 0.002\\\hline
Precisão & \textbf{0.981 $\pm$ 0.002} & 0.694 $\pm$ 0.434 & 0.975 $\pm$ 0.004\\\hline
Recall & \textbf{0.981 $\pm$ 0.001} & 0.531 $\pm$ 0.323 & 0.943 $\pm$ 0.005\\\hline
F1-Score & \textbf{0.981 $\pm$ 0.001} & 0.600 $\pm$ 0.369 & 0.959 $\pm$ 0.002\\\hline
\end{tabular}
}
}
\end{center}
\end{table}

\section{Conclusões}
Este trabalho propõe a utilização de algoritmos de aprendizado de máquina para a detecção de ataques em redes. Estes algoritmos recebem dados extraídos por um AE, treinado somente utilizando dados normais, isto é, sem ataque, dando origem a um ADS baseado em AE.
Para isso, alguns algoritmos de classificação foram analisados: o KNN, \textit{K-Means} e a SVM. Os resultados também foram comparados com o método de obtenção de um limiar desenvolvido em~\cite{Choi2019}.

Os resultados mostram que o algoritmo KNN teve um melhor desempenho comparado com a Referência em todos os cenários avaliados neste trabalho. O segundo melhor resultado foi obtido pela SVM, o terceiro foi a Referência e o pior foi ao utilizar o \textit{K-Means}. Uma possível explicação para os resultados obtidos com o \textit{K-Means}, pode ter relação com a convergência do algoritmo, que não foi atingida com os dados utilizados para seu treinamento, sendo esta uma hipótese que pode ser avaliada em trabalhos futuros.

Outro resultado importante foi que a utilização do RE em conjunto com a SIL, comparada somente com o uso da SIL, não proporcionou uma melhora dos resultados.

Também foi possível observar que utilizar somente a SIL traz um desempenho melhor do que utilizar somente o RE.

Com isso, analisando todos os resultados obtidos, nota-se que a utilização de um AE com o KNN para classificação da presença de um ataque na construção de um ADS, mostra-se muito eficiente comparado à utilização de somente um limiar sob o valor do RE obtido através de um AE.

Alguns trabalhos futuros poderão ser realizados com o foco na própria estrutura do AE para melhora dos resultados, alterando a quantidade de camadas, neurônios e parâmetros do AE.



\begin{thebibliography}{10}
\providecommand{\url}[1]{#1}
\csname url@samestyle\endcsname
\providecommand{\newblock}{\relax}
\providecommand{\bibinfo}[2]{#2}
\providecommand{\BIBentrySTDinterwordspacing}{\spaceskip=0pt\relax}
\providecommand{\BIBentryALTinterwordstretchfactor}{4}
\providecommand{\BIBentryALTinterwordspacing}{\spaceskip=\fontdimen2\font plus
\BIBentryALTinterwordstretchfactor\fontdimen3\font minus
  \fontdimen4\font\relax}
\providecommand{\BIBforeignlanguage}[2]{{%
\expandafter\ifx\csname l@#1\endcsname\relax
\typeout{** WARNING: IEEEtran.bst: No hyphenation pattern has been}%
\typeout{** loaded for the language `#1'. Using the pattern for}%
\typeout{** the default language instead.}%
\else
\language=\csname l@#1\endcsname
\fi
#2}}
\providecommand{\BIBdecl}{\relax}
\BIBdecl

\bibitem{Moustafa2019}
N.~Moustafa, J.~Hu, and J.~Slay, ``{A holistic review of Network Anomaly
  Detection Systems: A comprehensive survey},'' \emph{Journal of Network and
  Computer Applications}, vol. 128, pp. 33--55, 2019.

\bibitem{Ahmed2016}
M.~Ahmed, A.~{Naser Mahmood}, and J.~Hu, ``{A survey of network anomaly
  detection techniques},'' \emph{Journal of Network and Computer Applications},
  vol.~60, pp. 19--31, 2016.

\bibitem{Fernandes2019}
G.~Fernandes, J.~J. Rodrigues, L.~F. Carvalho, J.~F. Al-Muhtadi, and M.~L.
  Proen{\c{c}}a, ``{A comprehensive survey on network anomaly detection},''
  \emph{Telecommunication Systems}, vol.~70, pp. 447--489, 2019.

\bibitem{Kurniabudi2019}
Kurniabudi, B.~Purnama, Sharipuddin, Darmawijoyo, D.~Stiawan, Samsuryadi,
  A.~Heryanto, and R.~Budiarto, ``{Network anomaly detection research: A
  survey},'' \emph{Indonesian Journal of Electrical Engineering and
  Informatics}, vol.~7, pp. 36--49, 2019.

\bibitem{Manna2019}
A.~Manna and M.~Alkasassbeh, ``{Detecting network anomalies using machine
  learning and SNMP-MIB dataset with IP group},'' \emph{2019 2nd International
  Conference on New Trends in Computing Sciences, ICTCS 2019 - Proceedings},
  pp. 2--6, 2019.

\bibitem{Al-Naymat2018}
G.~Al-Naymat, M.~Al-Kasassbeh, and E.~Al-Hawari, ``Exploiting snmp-mib data to
  detect network anomalies using machine learning techniques,'' in
  \emph{Proceedings of SAI Intelligent Systems Conference}.\hskip 1em plus
  0.5em minus 0.4em\relax Springer, 2018, pp. 991--1004.

\bibitem{Toupas2019}
P.~Toupas, D.~Chamou, K.~M. Giannoutakis, A.~Drosou, and D.~Tzovaras, ``{An
  intrusion detection system for multi-class classification based on deep
  neural networks},'' \emph{Proceedings - 18th IEEE International Conference on
  Machine Learning and Applications, ICMLA 2019}, pp. 1253--1258, 2019.

\bibitem{Charte2018}
D.~Charte, F.~Charte, S.~Garc{\'{i}}a, M.~J. del Jesus, and F.~Herrera, ``{A
  practical tutorial on autoencoders for nonlinear feature fusion: Taxonomy,
  models, software and guidelines},'' \emph{Information Fusion}, vol.~44, pp.
  78--96, 2018.

\bibitem{Sakurada2014}
M.~Sakurada and T.~Yairi, ``{Anomaly detection using autoencoders with
  nonlinear dimensionality reduction},'' \emph{ACM International Conference
  Proceeding Series}, vol. 02-Decembe, pp. 4--11, 2014.

\bibitem{Madani2018}
P.~Madani and N.~Vlajic, ``{Robustness of deep autoencoder in intrusion
  detection under adversarial contamination},'' \emph{ACM International
  Conference Proceeding Series}, 2018.

\bibitem{Zavrak2020}
S.~Zavrak and M.~Iskefiyeli, ``{Anomaly-Based Intrusion Detection From Network
  Flow Features Using Variational Autoencoder},'' \emph{IEEE Access}, vol.~8,
  pp. 108\,346--108\,358, 2020.

\bibitem{Choi2019}
H.~Choi, M.~Kim, G.~Lee, and W.~Kim, ``{Unsupervised learning approach for
  network intrusion detection system using autoencoders},'' \emph{Journal of
  Supercomputing}, vol.~75, pp. 5597--5621, 2019.

\bibitem{Hwang1999}
B.~Hwang and S.~Cho, ``{Characteristics of autoassociative MLP as a novelty
  detector},'' \emph{Proceedings of the International Joint Conference on
  Neural Networks}, vol.~5, pp. 3086--3091, 1999.

\bibitem{Yu2017}
Y.~Yu, J.~Long, and Z.~Cai, ``{Network Intrusion Detection through Stacking
  Dilated Convolutional Autoencoders},'' \emph{Security and Communication
  Networks}, vol. 2017, pp. 1--10, 2017.

\bibitem{Kramer2013}
O.~Kramer, \emph{{Dimensionality Reduction with Unsupervised Nearest
  Neighbors}}.\hskip 1em plus 0.5em minus 0.4em\relax Springer, 2013, vol.~51.

\bibitem{Blomer2016}
J.~Bl{\"o}mer, C.~Lammersen, M.~Schmidt, and C.~Sohler, ``Theoretical analysis
  of the k-means algorithm--a survey,'' in \emph{Algorithm Engineering}.\hskip
  1em plus 0.5em minus 0.4em\relax Springer, 2016, pp. 81--116.

\bibitem{Yu2008}
J.~Yu, H.~Lee, M.-S. Kim, and D.~Park, ``{Traffic flooding attack detection
  with SNMP MIB using SVM},'' \emph{Computer Communications}, vol.~31, no.~17,
  pp. 4212--4219, 2008.

\bibitem{NSL-KDD}
``{NSL-KDD Dataset},'' \url{https://www.unb.ca/cic/datasets/nsl.html}, 2009
  (accessed May 30, 2022).

\bibitem{KDD99}
``{The Third International Knowledge Discovery and Data Mining Tools
  Competition},''
  \url{http://kdd.ics.uci.edu/databases/kddcup99/kddcup99.html}, 1999 (accessed
  May 30, 2022).

\end{thebibliography}



\end{document}